\documentclass[lettersize,journal]{IEEEtran}
\usepackage{amsmath,amsfonts,amssymb}
\usepackage{array}
\usepackage[caption=false,font=normalsize,labelfont=sf,textfont=sf]{subfig}
\usepackage{textcomp}
\usepackage{stfloats}
\usepackage{url}
\usepackage{verbatim}
\usepackage{graphicx}
\usepackage{cite}
\usepackage{hyperref}
\usepackage{float}
\usepackage{threeparttable}
\usepackage{multirow}
\usepackage{epsfig} 
\usepackage{epstopdf}
\usepackage{diagbox}
\usepackage{booktabs} 
\usepackage{algpseudocode}
\usepackage[table]{xcolor}
\usepackage{colortbl}
\usepackage{color}
\usepackage{makecell}
\usepackage{bbding}
\usepackage{caption}
\usepackage{pifont}

\definecolor{NavyBlue}{RGB}{0, 50, 200}

\begin{document}

\title{ToLL: Topological Layout Learning with Asymmetric Cross-View Structural Distillation for 3D Scene Graph Generation Pretraining}

\author{Yucheng~Huang$^*$,
        Luping~Ji, \emph{Member, IEEE}, Xiangwei Jiang$^*$, and Wen Li, Mao Ye
\thanks{This work was supported by the National Natural Science Foundation of China (NSFC) under Grant 62476049. $^*$ Equal contribution. (\emph{Corresponding author: Luping Ji}.)}%
\thanks{The authors are with the School of Computer Science and Engineering, University of Electronic Science and Technology of China, Chengdu 611731, China (email: \{hyc,jxw\}@std.uestc.edu.cn; \{jiluping,liwen,maoye\}@uestc.edu.cn).}}

\markboth{Journal of \LaTeX\ Class Files,~Vol.~14, No.~8, August~2021}%
{Shell \MakeLowercase{\textit{et al.}}: A Sample Article Using IEEEtran.cls for IEEE Journals}

\maketitle

\begin{abstract}
3D Scene Graph (3DSG) generation plays a pivotal role in spatial understanding and affordance perception. To mitigate generalization issues from data scarcity, joint-embedding and generative proxy tasks are proposed to pre-train 3DSG representations on predicate label-free datasets. Currently, generative pre-training usually bypasses the semantic corruption caused by the geometric augmentations in joint-embedding, but cannot avoid a negative problem ``Geometric Shortcut." In this problem, exposing dense object spatial and scale priors will induce models to trivially reconstruct scenes by interpolating object positions, rather than learning the underlying topological constraints provided by edges.
To address this issue, we propose a Topological Layout Learning (ToLL) for 3DSG generation pretraining framework. In detail, we design an Anchor-Conditioned Topological Geometry Reasoning. It adopts a recurrent GNN to recover the global layout of zero-centered subgraphs (the non-visible spatial features) by one anchor with sparse spatial prior. Considering the absence of spatial layout information within the objects, it creates an information bottleneck, compelling our model to recover the full scene layout by leveraging predicate representation learning. Moreover, we construct a Structural Multi-view Augmentation to avoid semantic corruption, enhancing 3DSG representations via self-distillation. The extensive experiments on special dataset demonstrate that our ToLL could often improve 3DSG pertaining quality, outperforming state-of-the-art baselines. Source codes are available at \href{https://github.com/UESTC-nnLab/ToLL-SGG}{https://github.com/UESTC-nnLab/ToLL-SGG.}
\end{abstract}


\section{Introduction}
\IEEEPARstart{T}{he} 3D Scene Graph (3DSG) generation elevates scene understanding from object-centric perception to holistic relational modeling. By capturing the semantic associations and functional affordances between entities, 3DSGs serve as a structured representation for downstream agents in tasks like Vision-Language Navigation~\cite{anderson2018vision} and Robotics~\cite{zitkovich2023rt}. A typical 3DSSG framework~\cite{wald2020learning,armeni20193d,zhang2021exploiting} comprises object and edge encoders for semantic extraction, coupled with a Graph Neural Network (GNN) for topological reasoning. Current 3DSG research primarily explores knowledge prior injection, predicate representation modeling, and self-supervised learning. 

\begin{figure}[t!]
	\centering
	\subfloat{
		\includegraphics[width=0.98\linewidth]{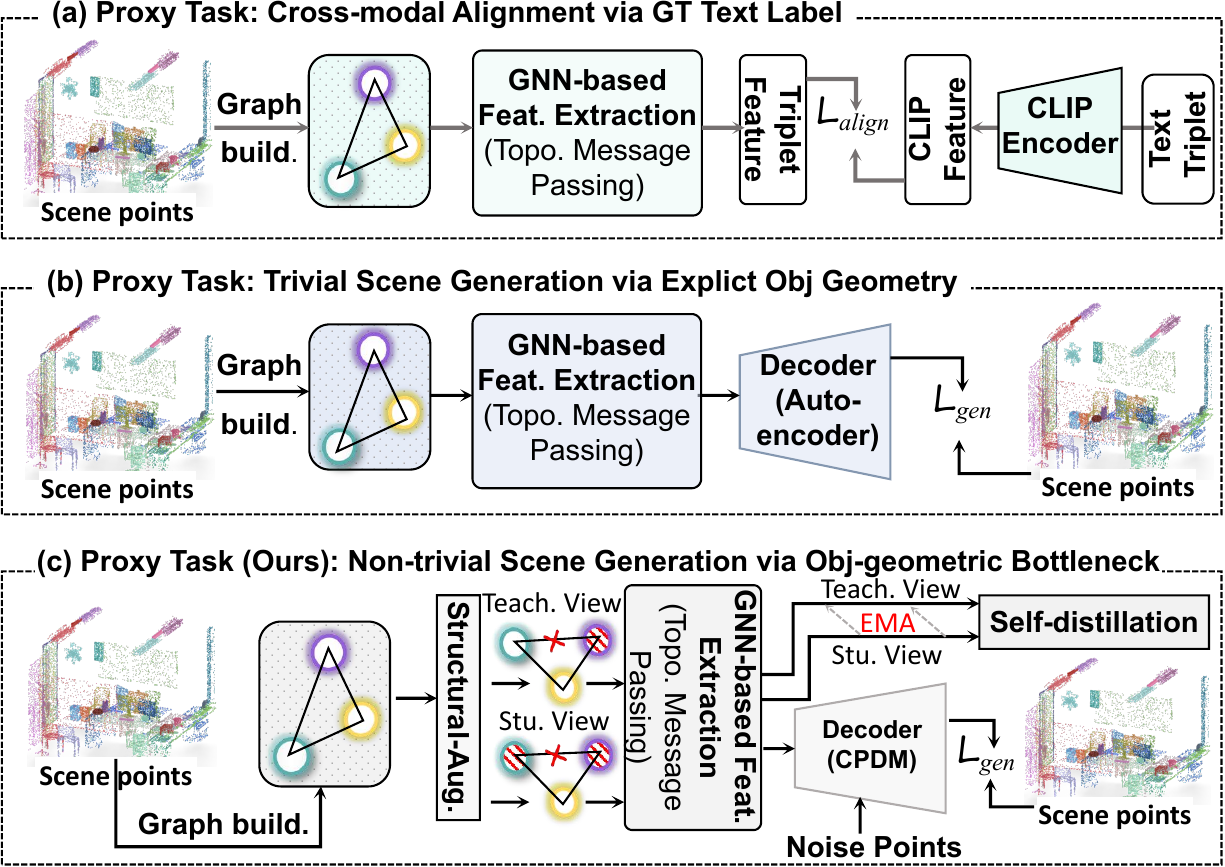}}
	\caption{Proxy task scheme comparison of 3DSG generation: (a) cross-modal Vision-Language representation,
    (b) trivial scene generation, 
    (c) non-trivial scene generation, $i.e.$, our ToLL.
    }
	\label{fig1} 
\end{figure}

Early approaches relied on knowledge prior injection, either through statistical prototypes~\cite{zhang2021knowledge} or cross-modal distillation from CLIP~\cite{wang2023vl,chen2024clip,koch2024lang3dsg} as shown in Figure~\ref{fig1}(a). However, these methods depend heavily on annotated text-label pairs, limiting scalability. Others focused on explicit geometric modeling, utilizing vector differences~\cite{wu2023incremental} or hyperrectangles~\cite{feng2025hyperrectangle} to constrain predicate learning. However, these supervised methods struggle with generalization due to the long-tail sparsity of 3D relational data.

To reduce annotation dependence, Self-Supervised Learning (SSL) has emerged as a promising direction. Joint-embedding methods~\cite{xie2020pointcontrast,wu2025sonata,zhang2021self,zhang2021self,van2025joint} have proven effective for point cloud understanding but falter in 3DSG, as predicate representations are sensitive to the geometric transformations required for multi-view augmentation. While MVIL~\cite{huang2026multi} decouples predicate learning via a two-stage training strategy to alleviate sensitivity to geometric transformations, its pre-training-induced rotation-invariant representations may compromise the downstream directional predicate learning. Additionally, the reliance of MVIL on advanced VLMs to produce semantic predicate relation pseudo-labels for the entire dataset leads to higher model training costs.

\begin{figure*}[t!]
	\centering
	\subfloat{
		\includegraphics[width=0.95\linewidth]{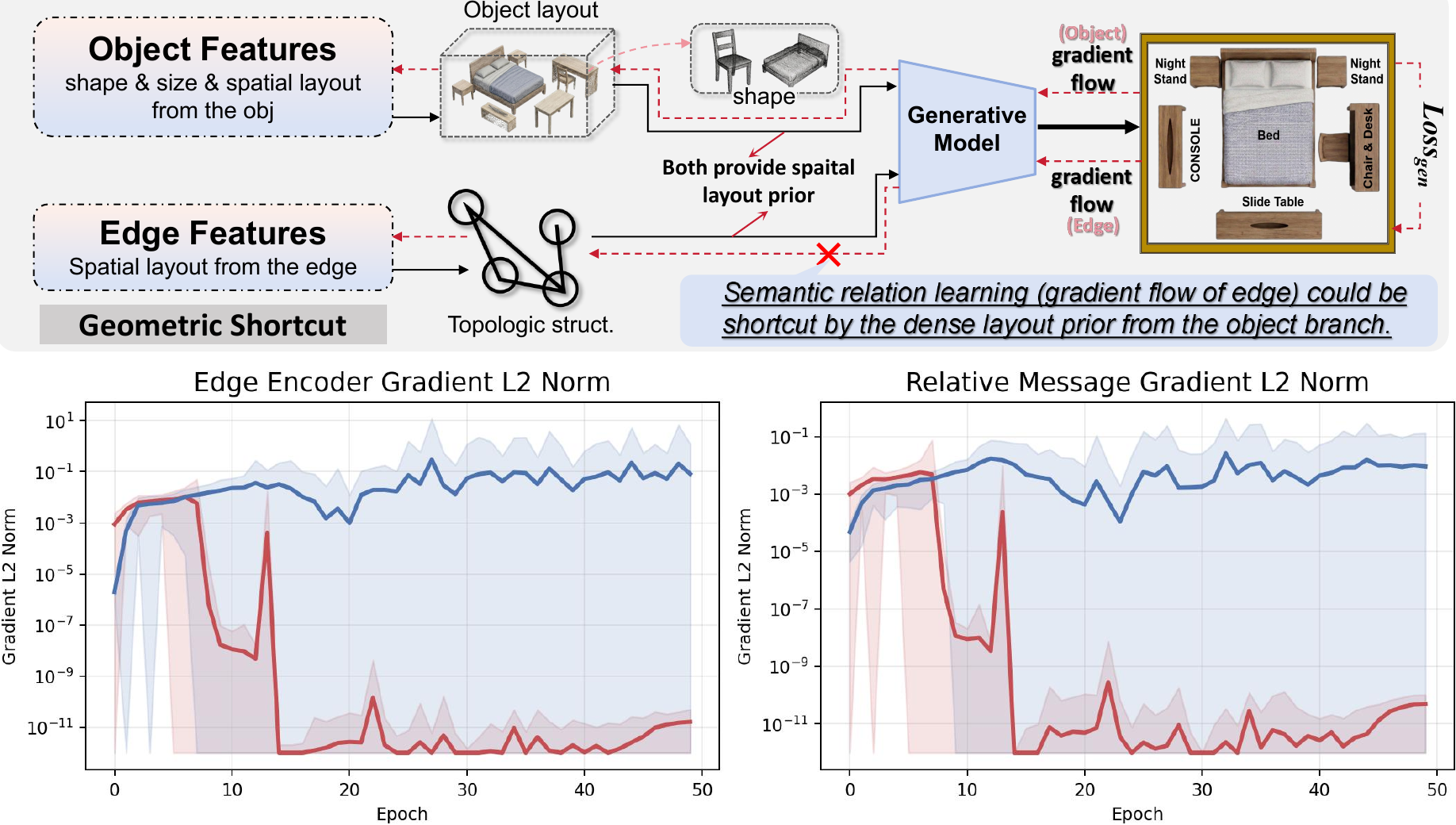}}
	\caption{The motivation of designing Anchor-Conditioned Topological Geometric Reasoning (ACTGR). We observe that exposing complete object geometries (shape, scale, and location) will trigger shortcut learning, causing edge encoder and relative message gradients (plotted by red curves) drop sharply. In contrast, ACTGR can effectively avoid this problem (refer to the blue curves).
    }
	\label{fig2} 
\end{figure*}

To bypass augmentation challenges, alternative approaches leverage generative pretext tasks for 3DSG representation learning. Inspired by Auto-Encoder methods~\cite{kingmaauto, van2017neural},~\cite{koch2024sgrec3d} designs an augmentation-free generative pretraining for 3DSG as shown in Figure~\ref{fig1}(b). However, we identify a critical flaw in existing generative graph approaches: the "Geometric Shortcut". As shown in Figure~\ref{fig2}, when dense spatial priors of objects are provided, models tend to trivially interpolate the node positions from neighbors or themselves, rather than learning the underlying topological constraints by edges.

To address these limitations, we propose a Topological Layout Learning (ToLL) framework as shown in Figure~\ref{fig1}(c). Our approach enforces the model to learn scene layout strictly from edge topology, preventing shortcut learning. First, we introduce Anchor-Conditioned Topological Geometric Reasoning (ACTGR). By normalizing object point clouds into a canonical space and applying Point-MAE~\cite{pang2023masked} style masking, we design a \textit{single-anchor information bottleneck} \cite{hu2024survey}. This mechanism compels the GNN to perform spatial "dead-reckoning": armed with the global spatial prior of only one anchor object, the model must recover the entire scene layout via edge-based topological propagation. This generative objective simultaneously enhances both intra-object geometric details and inter-object edge semantics.

To circumvent the over-smoothing issue in autoencoder-based geometric modeling, which degrades conditional semantic features, so we employ diffusion models to formulate the generative pretext task.

Second, to robustify representation against the semantic corruption from geometric transformation, we propose Structural Multi-view Augmentation (SMA) inspired by~\cite{caron2021emerging,wusimplifying,you2020graph,zhu2021graph,hou2022graphmae,zhouimage}. Instead of geometric distortion, we employ connectivity perturbation to generate asymmetric student views. Through a SwAV-style~\cite{caron2020unsupervised} self-distillation, we enforce semantic consistency, ensuring the backbone captures representations invariant to both spatial incompleteness and topological occlusion.

In summary, our primary contributions including: 

(1) We propose the ToLL, a new 3DSG pre-training framework that mitigates geometric shortcuts in the generative task. 

(2) We design ACTGR to build the information bottleneck for avoiding the edge learning shortcut by dense spatial prior from objects, and SMA to learn structurally robust semantics via cross-view self-distillation. 

(3) The extensive experiments are conducted on the public 3DSSG dataset to demonstrate that our ToLL scheme effectively boosts existing 3DSG generation methods.

\section{Related Work}

\textbf{3D Scene Graph Prediction.} 
Supervised 3DSG architectures~\cite{wald2020learning,ma2024heterogeneous,feng20233d, wang2024weakly, wei20233d} have evolved to incorporate external semantic priors (e.g., VLMs~\cite{koch2024lang3dsg,chen2024clip}) or explicit geometric constraints~\cite{wu2023incremental,feng2025hyperrectangle} for predicate refinement. However, these methods rely heavily on sparse annotated triplets, limiting generalization.

\textbf{Self-supervised Learning on Point Cloud.}
Self-Supervised Learning (SSL) on point clouds, including joint-embedding~\cite{xie2020pointcontrast,zhangconcerto} and masked modeling~\cite{pang2023masked}, addresses data scarcity but struggles in the 3DSG domain. Rigid geometric augmentations (e.g., rotation) used in SSL can fundamentally alter spatial predicate semantics. Furthermore, these object-centric methods~\cite{wu2025sonata} fail to capture the complex topological structures essential for inferring inter-object relationships. Despite~\cite{koch2024sgrec3d}'s success with generative pre-training, it ignores shortcuts arising from full node visibility, which undermine topological reasoning. This limitation prevents the model from acquiring effective predicate representation capabilities.

\textbf{Self-supervised Learning (SSL) on Graph.}
Graph SSL~\cite{liu2022graph,wu2021self} typically constructs multi-view pairs via topological augmentations—such as node dropping or edge perturbation employed in GraphCL~\cite{you2020graph,you2021graph,xu2021infogcl}, or through reconstructive objectives like masking node features and structures in GraphMAE~\cite{hou2022graphmae,tian2023heterogeneous}. Since 3DSG can be abstracted as a graph representation learning problem~\cite{hamilton2020graph}, these topological augmentations offer a crucial advantage. By constructing multiple structural views, we can facilitate robust representation learning while avoiding the semantic corruption of predicates often caused by the geometric augmentations.

\begin{figure*}[t!]
	\centering
	\subfloat{
		\includegraphics[width=0.95\linewidth]{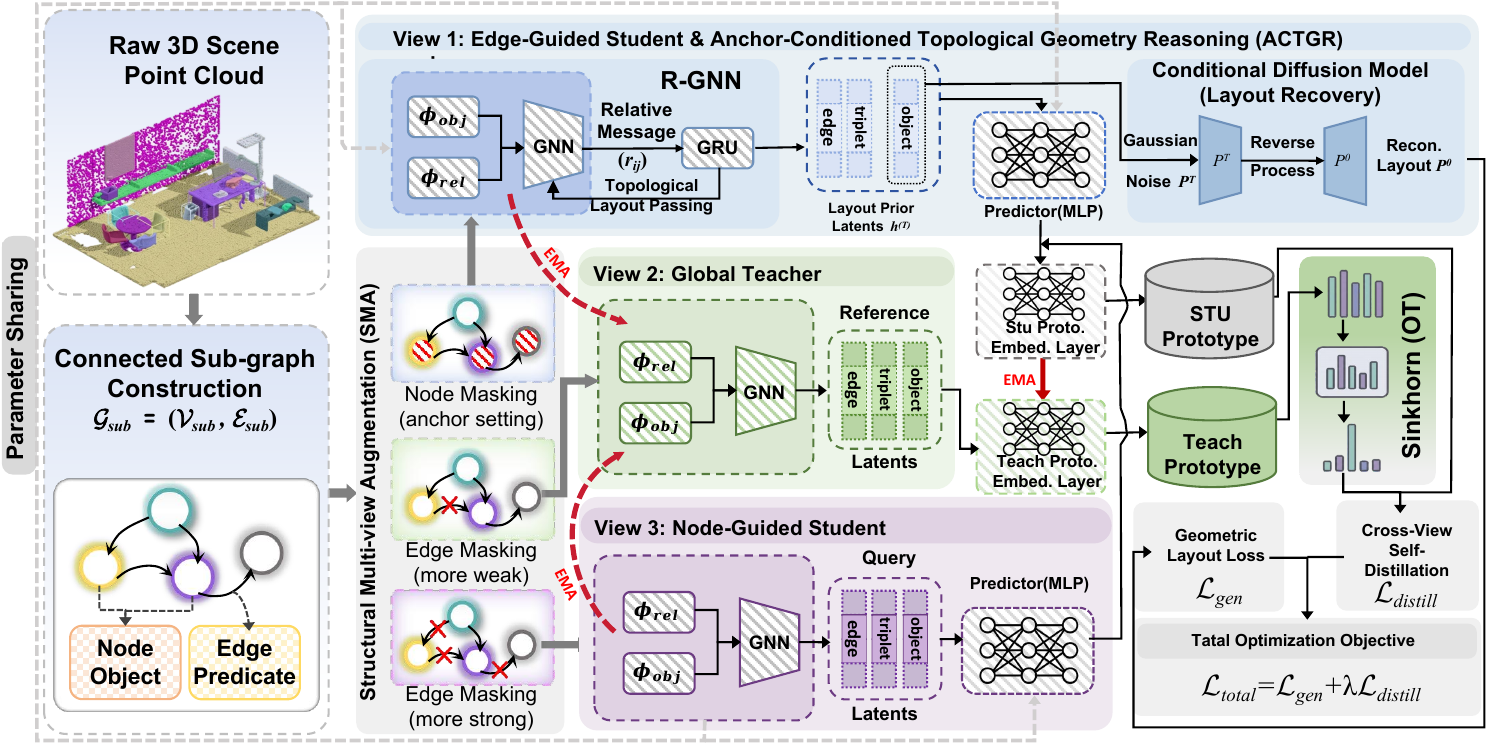}}
	\caption{Our 3D Scene Graph Pretraining scheme via Topological Layout Learning with Structural Multi-view Augmentation.}
	\label{fig3} 
\end{figure*}

\section{Preliminary}
\subsection{3D Scene Graph Generation}
\textbf{Definitions.} We formulate the task of 3D Scene Graph Generation as learning a mapping from a 3D scene point cloud to a semantic graph $\mathcal{G} = (\mathcal{V}, \mathcal{E})$. Here, $\mathcal{V} = \{o_i\}_{i=1}^N$ represents the set of $N$ object instances, where each $o_i$ is associated with a subset of point cloud data. The set of edges $\mathcal{E} = \{ e_{ij} \in \mathbb{R}^D \mid (i, j) \in \mathcal{M} \}$ denotes the semantic dependencies, where a directed edge $e_{ij} = (o_i, o_j)$ corresponds to a predicate relationship between the subject $o_i$ and object $o_j$. And $\mathcal{M}$ is the set of connected indices. The total number of edges is given by $|\mathcal{E}| = M$.

\textbf{Feature Encoding and Propagation.} The proposed framework comprises three core modules: an object encoder $\phi_{\text{obj}}$, a predicate encoder $\phi_{\text{rel}}$, and a graph neural network $\text{GNN}(\cdot)$. First, we initialize the latent embeddings for nodes and edges by projecting the raw point cloud data and relative geometric features into a high-dimensional latent space through
$
\{\mathbf{h}_i^{(0)} = \phi_{\text{obj}}(o_i), \quad \mathbf{h}_{ij}^{(0)} = \phi_{\text{rel}}(o_i, o_j)\}
$, 
where $\mathbf{h}_i^{(0)}\in \mathbb{R}^{N\times d}, \mathbf{h}_{ij}^{(0)} \in \mathbb{R}^{M\times d}$ denote the initial features for nodes and edges.

To capture high-order dependencies, these embeddings are propagated through a message-passing mechanism. The GNN aggregates context from neighbors, as follows:
\begin{equation}
\begin{aligned}
\{\mathbf{h}_i^{(L)}, \mathbf{h}_{ij}^{(L)}\} = \text{GNN}\left(\{\mathbf{h}_i^{(0)}\}, \{\mathbf{h}_{ij}^{(0)}\}; \mathcal{E}\right)
\end{aligned}
\label{equation1}
\end{equation}
where $\mathbf{h}^{(L)}$ represents the contextually refined representations after $L$ layers of propagation.

\textbf{Scene Graph Inference.} Finally, the refined embeddings are fed into task-specific classification heads (MLPs) to predict the semantic probability distributions
\begin{equation}
\begin{aligned}
P(c_i | \mathcal{G}) = \text{Softmax}(\text{MLP}_{\text{cls}}(\mathbf{h}_i^{(L)}))\\ \quad P(r_{ij} | \mathcal{G}) = \text{Softmax}(\text{MLP}_{\text{rel}}(\mathbf{h}_{ij}^{(L)}))
\end{aligned}
\label{equation1}
\end{equation}
where $c_i$ and $r_{ij}$ denote the predicted object class and predicate category, respectively.

\subsection{Conditional Point Cloud Diffusion}
We formulate the main pre-training task as a conditional generative process guided by latent codes $c$ with layout prior. Following the DDPM framework, we define a forward process that gradually corrupts the clean point cloud $X^0$ into Gaussian noise. At any step $t$, the noisy state $X^t$ can be sampled directly via $q(X^{t} | X^{0}) = \mathcal{N}(X^{t}; \sqrt{\bar{\alpha}_{t}}X^{0}, (1-\bar{\alpha}_{t})\mathbf{I})$, where $\bar{\alpha}_t$ is the cumulative noise schedule.

The generative model learns to reverse this process by predicting the noise $\epsilon$ added to $X^t$, conditioned on the latent representation $c$. We optimize the network parameters $\theta$ by minimizing the simple noise prediction error:
\begin{equation}
\mathcal{L}(\theta) = \mathbb{E}_{t, X^0, \epsilon} \left[ \left\| \epsilon - \epsilon_{\theta}(\sqrt{\bar{\alpha}_t} X^0 + \sqrt{1-\bar{\alpha}_t} \epsilon, c, t) \right\|^2 \right],
\end{equation}
where $t$ is uniformly sampled, $\epsilon \sim \mathcal{N}(0, I)$, $\epsilon_\theta(\cdot)$ is a trainable neural network. This objective forces the backbone to capture the underlying geometric priors of 3D objects under the structural guidance of $c$.

\section{Methods}
We mainly formulate 3D Scene Graph pre-training (Figure~\ref{fig3}) as a diffusion-based conditional layout restoration task. 

\subsection{Anchor-Conditioned Topological Geometric Reasoning.}
\textbf{Definitions.} Let $\mathcal{G}_{sub} = (\mathcal{V}_{sub}, \mathcal{E}_{sub})$ be a connected subgraph derived from the scene, where $\mathcal{V}_{sub}$ represents the object nodes and $\mathcal{E}_{sub}$ represents the relative geometric constraints (edges). Assuming the numbers of nodes and edges are $N$ and $M$, respectively. We define the problem of spatial layout restoration as inferring the geometric shape and the absolute spatial attributes $\mathbf{S} = \{\mathbf{s}_i\}_{i \in \mathcal{V}_{sub}}$ for all nodes, given a partially observed state driven by sparse anchors.

\textbf{Anchor Latent Initialization.} We first project all object point clouds $P_i$ within the subgraph $\mathcal{G}_{sub}$ into a canonical coordinate system via zero-mean centering and scale normalization. These normalized point clouds are processed by an object encoder $\phi_{\text{obj}}$ to extract latent features $\mathbf{h}_i^{(0)} \in \mathbb{R}^{N\times d}$. 

We define the spatial layout restoration as a conditional inference task anchored by a few nodes $\{v_a\} \in \mathcal{V}_{sub}$. For the selected anchor, we augment its latent feature with the 11-dimensional ground-truth absolute descriptor $\mathbf{s}_{\text{gt}}$, which encapsulates centroid coordinates, bounding box dimensions, and volumetric statistics. The final initial states $\mathbf{h}_i^{(0)}$ for the reasoning process are formulated as:
\begin{equation}
    \mathbf{h}_i'^{(0)} = 
    \begin{cases} 
        \text{MLP} \left( [\mathbf{h}_i^{(0)} \parallel \mathbf{s}_{\text{gt}}] \right), & i \in \mathcal{V}_{\text{anchor}} \\ 
        \mathbf{h}_i^{(0)}, & i \notin \mathcal{V}_{\text{anchor}}
    \end{cases}
    \label{eq:anchor_update}
\end{equation}
where $[\cdot \parallel \cdot]$ denotes the concatenation operation, and $\text{MLP}: \mathbb{R}^{d+11} \to \mathbb{R}^{d}$ is a Multi-Layer Perceptron (MLP) designed to project the concatenated features to the original latent dimension $d$, ensuring dimensional alignment.

\textbf{Relative Geometric Constraints.} Each edge $e_{ij} \in \mathcal{E}_{sub}$ carries a relative geometric attribute $\mathbf{r}_{ij}$ (e.g., relative vector $\Delta \mathbf{p}_{ij}$ or size ratio $\Delta \mathbf{s}_{ij}$). We posit that the absolute state of a neighbor $v_j$ can be recovered from $v_i$ via a learnable transformation function $\Phi_{\text{trans}}$:
\begin{equation}
\begin{aligned}
\mathbf{s}_j \approx \Phi_{\text{trans}}(\mathbf{s}_i, \mathbf{r}_{ij})
\end{aligned}
\label{equation1}
\end{equation}
if the graph is consistent, $\Phi_{\text{trans}}$ represents a geometric operator (e.g., translation or scaling) such that $\mathbf{s}_j = \mathbf{s}_i \oplus \mathbf{r}_{ij}$.

\textbf{Recurrent Topological Propagation.}
Since the depth of subgraphs varies and can be large, simply stacking GNN layers to cover the maximum possible path length is parameter-inefficient. To address this, we propose a recurrent propagation scheme. We utilize a lightweight, fixed-depth GNN (denoted as $\text{GNN}_{\text{base}}$, with $L_{\text{base}}=2$ layers) as the local spatial context aggregator, and wrap it within a Gated Recurrent Unit (GRU) to iteratively refine the node states.

At each recurrent step $t$ ($t=1, \dots, T$), the node features are updated by fusing the historical state with the newly aggregated local geometric messages:
\begin{equation}
\resizebox{0.95\columnwidth}{!}{$
    \mathbf{h}_i^{(t)} = \text{GRU} \bigg( \mathbf{h}_i^{(t-1)}, \underbrace{\text{GNN}_{\text{base}} \Big( \mathbf{h}_i^{(t-1)}, \sum_{j \in \mathcal{N}(i)} \psi_{\text{msg}} ( \mathbf{h}_j^{(t-1)}, \mathbf{e}_{ji}, \mathbf{r}_{ji} ) \Big)}_{\text{Spatial Context (2-layer GNN)}} \bigg)
$}
\label{eq:r_gnn_update}
\end{equation}
Here, $\psi_{\text{msg}}$ denotes the message function that encodes relative geometric constraints from neighbors, and $\text{GNN}_{\text{base}}$ aggregates these messages to form the local spatial context.

\textbf{Effective Receptive Field Analysis.}
Instead of fixing a deep architecture, our recurrent approach allows for a large and effective receptive field (ERF). With $T$ iterations and a base GNN depth of $L_{\text{base}}$, the ERF expands to $R_{\text{eff}} = T \times L_{\text{base}}$. For any target node $v_k$ with a shortest path distance $d(v_a, v_k) = K$ from the anchor, the absolute spatial information is fully recovered once the propagation covers the distance:
\begin{equation}
\begin{aligned}
\mathbf{h}_k^{(T)} \text{ encodes } \mathbf{s}_k \quad \text{if} \quad T \times L_{\text{base}} \ge K
\end{aligned}
\label{eq:reachability}
\end{equation}
This formulation effectively solves the chain of geometric constraints $\mathbf{s}_k = \mathbf{s}_a \oplus \mathbf{r}_{a,1} \oplus \dots \oplus \mathbf{r}_{K-1, k}$ in a parameter-efficient manner, as the model size remains constant regardless of the graph depth. Given a maximum subgraph depth does not exceed 10, we set $T=5$ to ensure full coverage.

\textbf{Latent-Guided Layout Recovery.}
Upon completion of the R-GNN propagation, we obtain latent representations $\mathcal{H} = \{\mathbf{h}_i^{(T)}\}_{i \in \mathcal{V}_{sub}}$. We cast the layout restoration as learning the conditional distribution $p_\theta(P | \mathcal{H})$ to reconstruct the scene of the subgraph. Since $\mathbf{h}_i^{(T)}$ has aggregated global topological messages relative to the anchor, it allows the generative model to recovery the scene point cloud with spatial layout. 

In layout learning, if there are more spatial information visible object anchors, edge networks may tend to be shortcut by explicit spatial priors from multiple objects, rather than learning the relative geometric constraints encoded in edges $\mathcal{E}_{sub}$. To strictly enforce topological reasoning, we propose a \textit{Single-Anchor Constraint Strategy.}

Formally, let $\mathcal{I}(\mathbf{S}; \mathcal{H})$ be the mutual information between the ground-truth spatial priors $\mathbf{S}$ from objects and the learned edge latent representations $\mathcal{H}$. Our goal is to maximize the dependency of $\mathcal{H}$ on the topological path $\mathcal{P}$ rather than on a dense set of absolute priors. We hypothesize that minimizing the number of anchors maximizes the "topological necessity."

\paragraph{Definition (Topological Necessity)} We strictly limit the anchor set to a singleton $|\mathcal{V}_{\text{anchor}}| = 1$. Let $v_a$ be the sole anchor. For any target node $v_k$, the information flow is forced to traverse the path $\mathcal{P}_{a \to k}$:
\begin{equation}
    p(\mathbf{s}_k | \mathbf{s}_a) \propto \prod_{e_{ij} \in \mathcal{P}_{a \to k}} \phi_{\text{edge}}(\mathbf{r}_{ij})
    \label{eq:single_anchor_prob}
\end{equation}
By masking all other anchors, we create an information bottleneck where recovering $\mathbf{s}_k$ is structurally impossible without decoding the relative transformations $\mathbf{r}_{ij}$ along the edges.

\paragraph{Proposition 1 (Prevention of Shortcut Learning)}
\textit{Let $\epsilon_{top}$ and $\epsilon_{sprior}$ denote error contributions from edge reasoning and the spatial prior provided by anchor. With multiple anchors ($|\mathcal{V}_{\text{anchor}}| \ge 2$), the model may trivially minimize loss via $\epsilon_{sprior}$ while neglecting $\epsilon_{top}$. Under the Single-Anchor Constraint ($|\mathcal{V}_{\text{anchor}}| = 1$), $\epsilon_{sprior}$ from objects with spatial information invisible becomes inaccessible, thereby forcing the optimization to minimize $\epsilon_{top}$.}

Empirically, we design a gradient analysis of edge modules as shown in Figure~\ref{fig2} to illustrate the   \textit{Prevention of Shortcut Learning}. And we randomly sample exactly one node $v_a \sim \mathcal{U}(\mathcal{V}_{sub})$ as the anchor per iteration. This compels the R-GNN to function as a rigorous "dead reckoning" system, ensuring learned features reflect cumulative geometric transformations rather than local memorization.

\subsection{Structural Multi-view Augmentation}
\label{sec:structural_aug}

To facilitate robust representation learning, we introduce \textbf{Structural Multi-view Augmentation (SMA)}. Unlike standard geometric augmentation, SMA employs topological perturbation and spatial masking to construct complementary views, enforcing semantic consistency via self-distillation.

We construct three different types of structural views that interact via a student-teacher paradigm in the following:

\textbf{Edge-Guided Student View, $\mathcal{G}_{edge}$:} This view preserves the almost complete \textbf{edge topology} but lacks absolute spatial attributes for the nodes (masking nodes). It aligns with the reference to learn geometry-to-semantics mapping.
    
\textbf{Holistic Teacher View, $\mathcal{G}_{ref}$:} A global view processed by the target teacher network. It retains both rich node information and stable edge topology, serving as the \textbf{comprehensive upper bound} for the students.
    
\textbf{Node-Guided Student View, $\mathcal{G}_{node}$:} This view preserves the \textbf{rich node information} but suffers from severe topological occlusion (masking edges). It forces the encoder to rely on node features to infer missing relationships. It aligns with the reference to learn semantics-to-topology reasoning.

\paragraph{Asymmetric Cross-View Distillation.}
We adopt a Mean Teacher framework consisting of a student network $f_\theta$ and a target network $f_\xi$. To prevent representational collapse, we employ an asymmetric architecture where a predictor head $g_\phi$ is appended exclusively to the student branch.

The target network parameters $\xi$ are updated via Exponential Moving Average (EMA) of the student parameters $\theta$:
\begin{equation}
    \xi_t = \alpha \xi_{t-1} + (1 - \alpha) \theta_t
\end{equation}
where $\alpha$ is the momentum coefficient.

\paragraph{View Settings.}
As shown in Table~\ref{tab:view_settings}, we generate three views for the Student network ($S_{v1}, S_{v2}, S_{v3}$) and two views for the Teacher network ($T_{v4}, T_{v5}$). 

The augmentation strategies include: 

\textbf{No Rotation Augmentation:} Denoted as ``Augmented'' in ~\ref{tab:view_settings}. This involves scene scale variations, random resampling and elastic distortion applied to the point cloud inputs.
    
    \textbf{Point Masking:} Randomly masking a portion of input point groups like Point-MAE.
    
    \textbf{Edge Masking:} Randomly masking edges to simulate partial connectivity. To maintain physical connectivity, learnable tensors are employed to replace the masked edges.

The Student network is trained to predict the semantic cluster assignments of the Teacher's views. Specifically, we enforce cross-view consistency pairs: $(T_{v4} \to S_{v2})$, $(T_{v5} \to S_{v3})$, $(T_{v5} \to S_{v1})$, and $(T_{v4} \to S_{v3})$.

\begin{table}[h]
    \centering
    \caption{View settings. \textbf{Source Input} indicates whether the input points are from the original data or augmented ones. \textbf{M-Ratio} denotes the masking ratio. The ACTGR strategy is exclusively implemented on the $S_{v1}$ view.}
    \label{tab:view_settings}
    \setlength{\tabcolsep}{2pt}
    \resizebox{\linewidth}{!}{
    \begin{tabular}{l|c|c|c|c|c}
        \toprule
        \textbf{Role} & \textbf{View ID} & \textbf{Source Input} & \textbf{Point M-Ratio} & \textbf{Edge M-Ratio} & \textbf{Objective} \\
        \midrule
        \multirow{3}{*}{Student} 
        & $S_{v1}$ & Origin & 0.8 & 0.2 & \multirow{3}{*}{Prediction / Gradient Update} \\
        & $S_{v2}$ & Augmented & 0.8 & 0.6 & \\
        & $S_{v3}$ & Origin & 0.8 & 0.6 & \\
        \midrule
        \multirow{2}{*}{Teacher} 
        & $T_{v4}$ & Origin & 0.2 & 0.2 & \multirow{2}{*}{Target Generation (EMA)} \\
        & $T_{v5}$ & Augmented & 0.1 & 0.1 & \\
        \bottomrule
    \end{tabular}
    }
\end{table}

\subsection{Optimization Objectives}
\label{sec:optimization}

\paragraph{Decoupled Geometric Layout Restoration}
In our default version, we directly employ the diffusion model to restore the scene point clouds with spatial layout information. In this setting, the raw point clouds $P_i$ retain their absolute spatial coordinates and scales, thereby implicitly preserving the relative spatial relationships and size variations among objects. The default generative objective is computed by
\begin{equation}
    \mathcal{L}_{default} = \mathbb{E}_{\tau, P, \epsilon} \Big[ \sum_{i \in \mathcal{V}_{sub}} \| \epsilon - \epsilon_\theta(P_{i}^{\tau}, \tau, h_i^{(T)}) \|_F^2 \Big]
\end{equation}
where $\tau$ denotes the diffusion timestep, $T$ represents the total number of recurrent iterations in the R-GNN.

However, this default diffusion process suffers from a scale-variance gradient bias. Because the optimization operates on absolute coordinates, large objects tend to dominate the loss landscape, overwhelming the geometric and semantic learning of minor entities.

To mitigate this issue, we introduce an advanced decoupled version, dubbed \textbf{ToLL++}. Instead of predicting noise in the absolute space, ToLL++ decouples the generative objective $\mathcal{L}_{gen}$ by disentangling an object's intrinsic shape from its absolute spatial attributes. Specifically, for each object $o_i$, we decompose its raw point cloud $P_i$ into a canonical shape $P_i^{can}$, a bounding box scale $s_i \in \mathbb{R}^3$, and a spatial centroid $c_i \in \mathbb{R}^3$, defined by $P_i^{can} = (P_i - c_i) / s_i$. Consequently, the generative task is reformulated: the diffusion model $\epsilon_\theta$ is restricted to predicting noise residuals solely within the normalized canonical space, while two auxiliary MLPs ($\Phi_{size}$ and $\Phi_{loc}$) explicitly regress the scale and centroid from the R-GNN latent state $h_i^{(T)}$. The decoupled losses for ToLL++ are formulated as:
\begin{equation}
\left\{
\begin{aligned}
    \mathcal{L}_{shape} &= \mathbb{E}_{\tau, P^{can}, \epsilon} \Big[ \sum_{i \in \mathcal{V}_{sub}} \| \epsilon - \epsilon_\theta(P_{i}^{\tau, can}, \tau, h_i^{(T)}) \|_F^2 \Big] \\
    \mathcal{L}_{size} &= \sum_{i \in \mathcal{V}_{sub}} \| s_i - \Phi_{size}(h_i^{(T)}) \|_2^2 \\
    \mathcal{L}_{loc} &= \sum_{i \in \mathcal{V}_{sub}} \| c_i - \Phi_{loc}(h_i^{(T)}) \|_2^2
\end{aligned}
\right.
\end{equation}
The overall layout restoration objective of ToLL++ is the weighted sum of these components:
\begin{equation}
    \mathcal{L}_{gen} = \mathcal{L}_{shape} + \lambda_{size} \mathcal{L}_{size} + \lambda_{loc} \mathcal{L}_{loc}
\end{equation}

\paragraph{Structural Distillation Loss}
To learn robust topological semantics, we implement a structural self-distillation task within the SMA module. As shown in Figure~\ref{fig3}, features from the student branch ($\mathbf{z}_{stu}$) and teacher branch ($\mathbf{z}_{ref}$) are projected into a shared semantic space via \textit{Stu.} and \textit{Teach. prototype embedding layers}. We maintain dynamic memory banks (\textit{Stu./Teach. Prototype}) to store historical cluster centroids. The \textit{Sinkhorn (OT)} algorithm is then employed to compute an optimal transport plan, which serves as a pseudo-label to guide the student's online clustering. The distillation loss is defined as:
\begin{equation}
\mathcal{L}_{distill} = \sum_{u \in \{edge, node\}} \sum_{l \in \mathcal{L}^*} \ell_{\text{swav}} \big( g_\phi^l(\mathbf{z}_{\mathcal{G}_u}^{(l)}), \text{sg}(\mathbf{z}_{ref}^{(l)}); \mathbf{C}^l \big)
\label{eq:distill_loss}
\end{equation}
where $\mathcal{L}^*$ denotes the $\{object, edge, triplet\}$. $\mathbf{z}_{\mathcal{G}_u}^{(l)}$ denotes features from the edge-guided ($\mathcal{G}_{edge}$) or node-guided ($\mathcal{G}_{node}$) student views. $g_\phi$ is a predictor head to prevent collapse, and $\text{sg}(\cdot)$ denotes the stop-gradient operation on the teacher branch, which is updated via EMA.

\paragraph{Total Objective}
The final objective balances the decoupled generative task with the semantic alignment:
\begin{equation}
    \mathcal{L}_{total} = \mathcal{L}_{gen}(\mathcal{G}_{edge}) + \lambda \sum_{v \in \{edge, node\}} \mathcal{L}_{distill}(\mathcal{G}_{v}, \mathcal{G}_{ref})
\end{equation}
where $\lambda$ governs the regularization strength. This formulation ensures the shared backbone simultaneously masters accurate layout recovery via edge-guided view $\mathcal{G}_{edge}$ and robust semantic topology via $\mathcal{G}_{node}$.

\begin{table*}[ht]
\centering
\caption{Comparisons with state-of-the-arts on the 3DSSG dataset. The inference model backbones are denoted by symbols: ``\textit{pn}"~for PointNet and ``\textit{pt}"~for PointTransformer. ``\textit{PointDif}": Weights of object encoder initialized with PointDif~\cite{zheng2024point}. ``\textit{ToLL}" or ``\textit{ToLL++}": Weights of complete 3DSG encoders initialized with our ToLL. ``\textit{+MLP only}": Only MLP heads are fine-tuned.}
\label{tab1}
\resizebox{\textwidth}{!}{%
\begin{tabular}{ll cc cccc cc cc cc}
\toprule
\multirow{2}{*}{\textbf{Baselines}} & \multirow{2}{*}{\textbf{Pre-training Method}} & \multicolumn{2}{c}{\textbf{Object}} & \multicolumn{4}{c}{\textbf{Predicate}} & \multicolumn{2}{c}{\textbf{Triplet}} & \multicolumn{2}{c}{\textbf{SGCLs}} & \multicolumn{2}{c}{\textbf{PredCLs}} \\
\cmidrule(lr){3-4} \cmidrule(lr){5-8} \cmidrule(lr){9-10} \cmidrule(lr){11-12} \cmidrule(lr){13-14}
 & & A@1 & A@5 & A@1 & A@3 & mA@1 & mA@3 & mA@50 & mA@100 & mR@20 & mR@50 & mR@20 & mR@50 \\
\midrule

SGPN\textsubscript{\textit{pn}} \cite{wald2020learning} & $\times$ & 50.32 & 74.56 & 89.89 & 98.15 & 40.63 & 63.41 & 52.74 & 65.58 & 19.7 & 22.6 & 32.1 & 38.4 \\
SGFN\textsubscript{\textit{pn}} \cite{wu2023incremental}  & $\times$ & 53.67 & 77.18 & 90.19 & 98.17 & 41.89 & 70.82 & 58.37 & 67.61 & 20.5 & 23.1 & 46.1 & 54.8 \\
SGFN\textsubscript{\textit{pt}} & $\times$ & 56.04 & 79.37 & 89.22 & 97.65 & 46.69 & 71.84 & 60.05 & 70.26 & 29.3 & 30.8 & 55.7 & 62.1 \\
VL-SAT\textsubscript{\textit{pn}}~\cite{wang2023vl}  & $\times$ & 55.66 & 78.66 & 89.81 & 98.45 & 54.03 &  77.67 & 65.09 & 73.59 & 31.8 & 32.4 & 57.8 & 64.2 \\
VL-SAT\textsubscript{\textit{pt}} & $\times$ & 57.84 & 78.69 & 89.76 & 98.13 & 52.43 & 73.35 & 63.55 & 72.48 & 30.3 & 32.8 & 54.7 & 63.8 \\
CCL-3DSG\textsubscript{\textit{pn}}~\cite{chen2024clip}  & $\times$ & - & - & - & - & - & - & - & - & 35.0 & 37.3 & 59.1 & 66.7 \\
SGFN\textsubscript{\textit{pt}} & $\times$ (only MLP) & 40.88 & 71.28 & 78.56 & 91.63 & 19.85 & 42.79 & 39.97 & 49.11 & 19.5 & 24.8 & 35.5 & 39.7 \\
VL-SAT\textsubscript{\textit{pt}} & $\times$ (only MLP) & 38.62 & 71.15 & 81.27 & 92.89 & 22.74 & 43.96 & 40.13 & 51.03 & 19.8 & 27.5 & 35.2 & 40.5 \\
\midrule
SGFN\textsubscript{\textit{pt}} & PointDif~\cite{zheng2024point}  & 57.10 & 79.56 & 89.75 & 98.12 & 48.82 & 73.29 & 63.61 & 72.89 & 32.2 & 33.5 & 56.4 & 62.6 \\
VL-SAT\textsubscript{\textit{pt}} & PointDif & 58.54 & 79.93 & 89.74 & 98.06 & 52.66 & 73.33 & 64.86 & 73.66 & 33.5 & 34.2 & 58.4 & 64.8 \\

\midrule
SGFN\textsubscript{\textit{pt}}  &OCRL~\cite{heoobject}   & 57.43 & 80.07 & 90.11 & 98.21 & 50.63 & 75.68 & 63.79 & 73.48 & 31.7 & 33.8 & 59.6 & 65.2 \\
VL-SAT\textsubscript{\textit{pt}}             &OCRL  & 59.27 & 80.42 & 89.97 & 98.16 & 54.26 & 75.68 & 64.37 & 74.26 & 34.6 & 37.1 & 58.7 & 66.4\\

\midrule
VL-SAT\textsubscript{\textit{pt}} & MvIL~\cite{huang2026multi}  & 58.34 & 80.26 & 91.03 & \textbf{98.96} & \textbf{58.43} & 79.63 & \textbf{68.57} & 76.89 & 36.7 & 38.2 & 59.8 & 68.3 \\
VL-SAT\textsubscript{\textit{pn}} & MvIL & 56.87 & 79.77 & 90.86 & 98.65 & 56.84 & 76.21 & 66.67 & 74.93 & 33.6 & 34.2 & 57.4 & 65.6 \\

\midrule
\textbf{SGFN}\textsubscript{\textit{pt}} & \textcolor{NavyBlue}{\textit{ToLL}} & 58.68 & 80.62 & 90.42 & 98.53 & 54.59 & 81.36 & 66.58 & 74.32 & 35.2 & 36.6 & 58.7 & 66.3 \\
\textbf{VL-SAT}\textsubscript{\textit{pt}} & \textcolor{NavyBlue}{\textit{ToLL}} & 58.72 & 80.54 & 90.88 & 98.64 & 56.67 & 79.03 & 66.42 & 75.59 & 35.7 & 36.7 & 59.2 & 67.6 \\
\textbf{SGFN}\textsubscript{\textit{pt}} & \textcolor{NavyBlue}{\textit{Decoupled ToLL++}} & 60.64 & 80.97 & \textbf{91.24} & 98.73 & 56.19 & 80.79 & 67.85 & 76.25 & 36.1 & 38.3 & {59.8} & {69.1} \\
\textbf{VL-SAT}\textsubscript{\textit{pt}} & \textcolor{NavyBlue}{\textit{Decoupled ToLL++}} & \textbf{61.43} & \textbf{81.74} & 90.67 & 98.86 & 57.94 & \textbf{82.06} & 68.42 & \textbf{78.42} & \textbf{37.4} & \textbf{40.2} & \textbf{60.2} & \textbf{69.4} \\
\midrule
SGFN\textsubscript{\textit{pt}} & \textcolor{NavyBlue}{\textit{ToLL}} (only MLP) & 52.54 & 76.64 & 82.51 & 94.74 & 28.77 & 49.30 & 48.87 & 58.29 & 27.6 & 29.7 & 39.0 & 44.2 \\
VL-SAT\textsubscript{\textit{pt}} & \textcolor{NavyBlue}{\textit{ToLL}} (only MLP) & 54.41 & 77.23 & 85.59 & 95.18 & 32.16 & 50.84 & 50.05 & 60.65 & 30.3 & 31.2 & 46.7 & 57.4 \\
\bottomrule
\end{tabular}%
}
\end{table*}

\begin{table*}[ht]
\centering
\caption{Performance on Predicate Classification (Head, Body, Tail) and Triplet (Zero-Shot setting). ``\textit{pn}"~for PointNet and ``\textit{pt}"~for PointTransformer. ``\textit{ToLL}" or or ``\textit{ToLL++}": Weights of complete 3DSG backbone initialized with our ToLL.}
\label{tab2}
\resizebox{\textwidth}{!}{%
\begin{tabular}{l c ccccc}
\toprule
\multirow{2}{*}{\textbf{Method}} & \multirow{2}{*}{\textbf{Pre-training Method}} & \multicolumn{3}{c}{\textbf{Predicate Classification Metrics}} & \multicolumn{2}{c}{\textbf{Triplet Classification Metrics}} \\
\cmidrule(lr){3-5} \cmidrule(lr){6-7}
 & & Head (mA@3 / mA@5) & Body (mA@3 / mA@5) & Tail (mA@3 / mA@5) & Unseen (A@50 / A@100) & Seen (A@50 / A@100) \\
\midrule
SGFN\textsubscript{\textit{pn}}~\cite{wu2023incremental}          & $\times$ & 95.08 / 99.38 & 70.02 / 87.81 & 38.67 / 58.21 & 22.59 / 35.68 & 71.44 / 80.11 \\
SGFN\textsubscript{\textit{pt}}                        & $\times$ & 96.18 / 99.46 & 76.74 / 87.41 & 48.25 / 62.73 & 25.54 / 46.69 & 72.87 / 80.08 \\
VL-SAT\textsubscript{\textit{pn}}~\cite{wang2023vl}       & $\times$ & 96.31 / 99.21 & 80.03 / 93.64 & 52.38 / 66.13 & 31.28 / 47.26 & 75.09 / 82.25 \\
VL-SAT\textsubscript{\textit{pt}}                      & $\times$ & 95.97 / 98.92 & 78.75 / 88.84 & 54.44 / 68.90 & 32.87 / 46.32 & 77.28 / 83.37 \\
CCL-3DSG\textsubscript{\textit{pn}}~\cite{chen2024clip} & $\times$ & 98.54 / \textbf{99.78} & 84.72 / 96.03 & 61.24 / 75.91 & 36.72 / 52.47 & 80.58 / 88.92 \\
VL-SAT\textsubscript{\textit{pt}}  & OCRL~\cite{heoobject} & 96.29 / 98.96 & 81.85 / 94.03 & 58.74 / 65.83 & 33.97 / 46.82 & 78.91 / 85.87 \\
VL-SAT\textsubscript{\textit{pt}}  & MvIL~\cite{huang2026multi} & \textbf{98.67} / 99.53 & \textbf{86.25} / 95.36 & 63.42 / 76.65 & 39.75 / 55.83 & \textbf{83.26} / 88.37 \\

\midrule
SGFN\textsubscript{\textit{pt}}                        & \textcolor{NavyBlue}{\textit{ToLL~~~~}} & 96.68 / 99.12 & 82.29 / 93.60 & 57.45 / 72.31 & 35.18 / 48.94 & 77.85 / 84.42 \\
VL-SAT\textsubscript{\textit{pt}}                       & \textcolor{NavyBlue}{\textit{ToLL~~~~}} & 96.97 / 99.65 & 84.41 / 97.83 & 59.92 / \textbf{78.92} & 38.64 / 56.19 & 78.82 / \textbf{90.06} \\
SGFN\textsubscript{\textit{pt}}                        & \textcolor{NavyBlue}{\textit{ToLL++}} & 96.75 / 99.56 & 84.61 / 94.73 & 60.62 / 76.48 & 36.27 / 50.63 & 79.74 / 87.53 \\
VL-SAT\textsubscript{\textit{pt}}                       & \textcolor{NavyBlue}{\textit{ToLL++}} & 98.52 / 99.64 & 85.72 / \textbf{98.45} & \textbf{63.95} / 77.83 & \textbf{40.28} / \textbf{56.42} & 82.29 / 89.87 \\
\bottomrule
\end{tabular}%
}
\end{table*}

\section{Experiments}
\subsection{Pretraining Setup}
For pre-training, we constructed subgraph samples using 1,513 ScanNet scenes. We first excluded objects with fewer than 512 points. The remaining objects were uniformly sampled to 1,024 points. Furthermore, we construct connected subgraphs to generate 7,392 samples, which collectively contain 33,949 nodes and 61,599 edges.

The pre-training stage consists of 300 epochs. The AdamW optimizer is utilized with a weight decay of $10^{-4}$ and a learning rate of $10^{-3}$, modulated by a cosine scheduler with a 5-epoch warm-up. A weight $\lambda$ of 0.1 is assigned to the self-distillation loss component. The batchsize is set to 32. The experiments are performed on four RTX 3090 GPUs.

\subsection{Fine-tuning Setup} 
3DSSG dataset is employed as the benchmark for scene graph fine-tuning. we used the standard configuration of 160 object categories and 27 predicate categories. The optimizer and scheduler configurations remained consistent with the pre-training phase, except that the warm-up period was omitted and the batch size is set to 8. The top-k accuracy ``A@k" is employed to evaluate object and predicate classification. The mean top-k accuracy ``mA@k" was used to assess the impact of long-tail categories in predicates. Scene Graph Classification (SGCLs, with graph-constrained) and Predicate Classification (PredCls) are introduced to evaluate predicate recall capability - both employing the Top-k recall ``R@k".

We introduce two evaluation protocols to test ToLL:\\
\textbf{The 1st protocol is a full fine-tuning scheme.} In this setting, the parameters of the complete pretrained encoders are updatable; however, we assigned them a lower learning rate of $0.5 \times 10^{-4}$, while the MLP layers are set to $10^{-4}$.
\\
\textbf{The 2nd protocol is an MLP-only fine-tuning scheme.} The entire pre-trained encoders are frozen, and only the MLP layers are fine-tuned with a learning rate, $10^{-4}$.

\subsection{Comparison with State-of-the-Art Methods} 

\textbf{Quantitative Analysis.} Table~\ref{tab1} presents a comparative evaluation on the 3DSSG dataset across multiple paradigms: learning from scratch, initialization with a pre-trained object encoder (PointDif), previous 3DSG pre-training frameworks (OCRL~\cite{heoobject}, MvIL~\cite{huang2026multi}), and our proposed ToLL.

First, compared to the baseline trained from scratch (``VL-SAT\textsubscript{\textit{pt}}"), ``VL-SAT\textsubscript{\textit{pt}}+\textit{PointDif}" achieves gains of 0.70 and 0.23 on the Object ``A@1" and Predicate ``mA@1" metrics, respectively. This substantiates that enhanced object representations, acting as reliable semantic anchors, can bolster predicate classification capabilities via GNNs.

Crucially, our advanced framework ``VL-SAT\textsubscript{\textit{pt}} w/ \textit{Decoupled ToLL++}" significantly outperforms the from-scratch baseline and other pre-training methods, achieving state-of-the-art results across multiple metrics. Compared to the baseline ``VL-SAT\textsubscript{\textit{pt}}", we observe obvious gains of 5.51 in Predicate ``mA@1", 8.71 in Predicate ``mA@3", and 4.87 in Triplet ``mA@50". Furthermore, it consistently surpasses recent strong pre-training baselines like OCRL and MvIL in Object ``A@1" (61.43 vs. 58.34 for MvIL) and Scene Graph Generation metrics (e.g., SGCLs ``mR@50" reaches 40.2). 

Finally, the ``\textit{MLP Only}" experiment for ``VL-SAT\textsubscript{\textit{pt}}" reveals that while our frozen backbone initialized with ToLL supports robust object classification (``A@1": 54.41, outperforming the 38.62 of the from-scratch MLP-only baseline).

\textbf{Long-tail Robustness and Zero-Shot Analysis.} To further probe the robustness of our pre-training framework, we detail performance across varying predicate frequency distributions (Head, Body, Tail) and evaluate zero-shot generalization capabilities on unseen triplets.

\begin{figure}[h]
	\centering
	\subfloat{
		\includegraphics[width=0.95\linewidth]{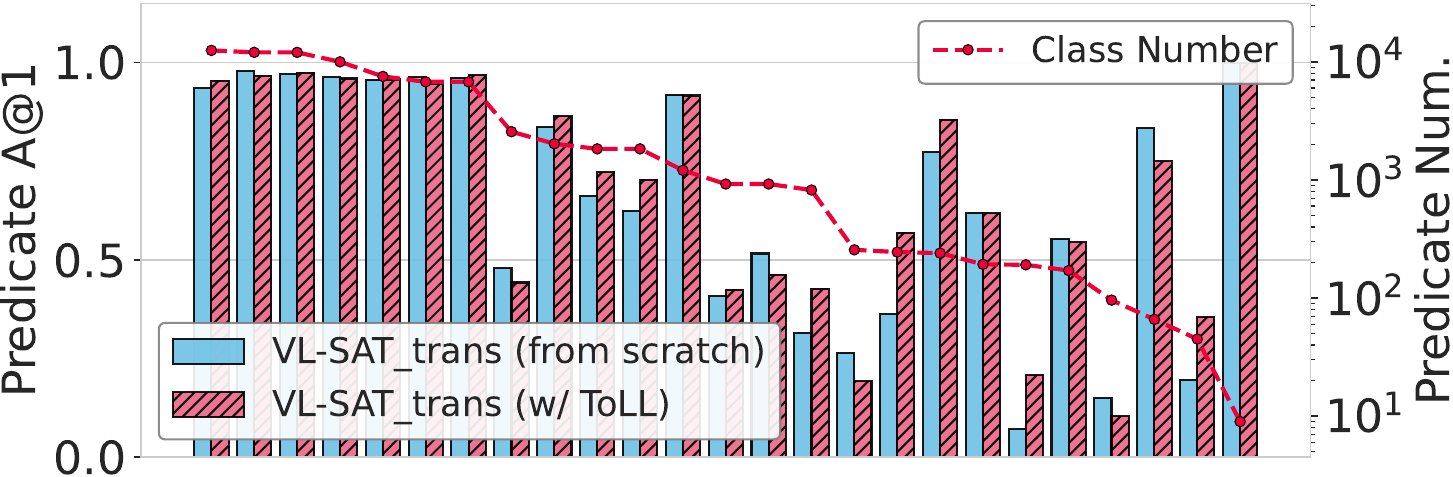}}
	\caption{Predicate A@1 for all predicate categories.}
	\label{fig4} 
\end{figure}

\textit{Long-tail Robustness.} As shown in Table~\ref{tab2}, while maintaining competitive performance on Head predicates, our method achieves remarkable gains on the long-tail categories. Specifically, ``VL-SAT\textsubscript{\textit{pt}} w/ \textit{ToLL++}" improves Body ``mA@3" by 6.97 (85.72 vs. 78.75) and surges Tail ``mA@3" by 9.51 (63.95 vs. 54.44) compared to the from-scratch baseline. It also demonstrates competitive long-tail mitigation compared to the recent MvIL. As shown in Figure~\ref{fig4}, leveraging our pre-trained weights enables the model to achieve higher recognition accuracy on long-tail predicate classes.

\begin{figure*}[h]
	\centering
	\subfloat{
		\includegraphics[width=0.98\linewidth]{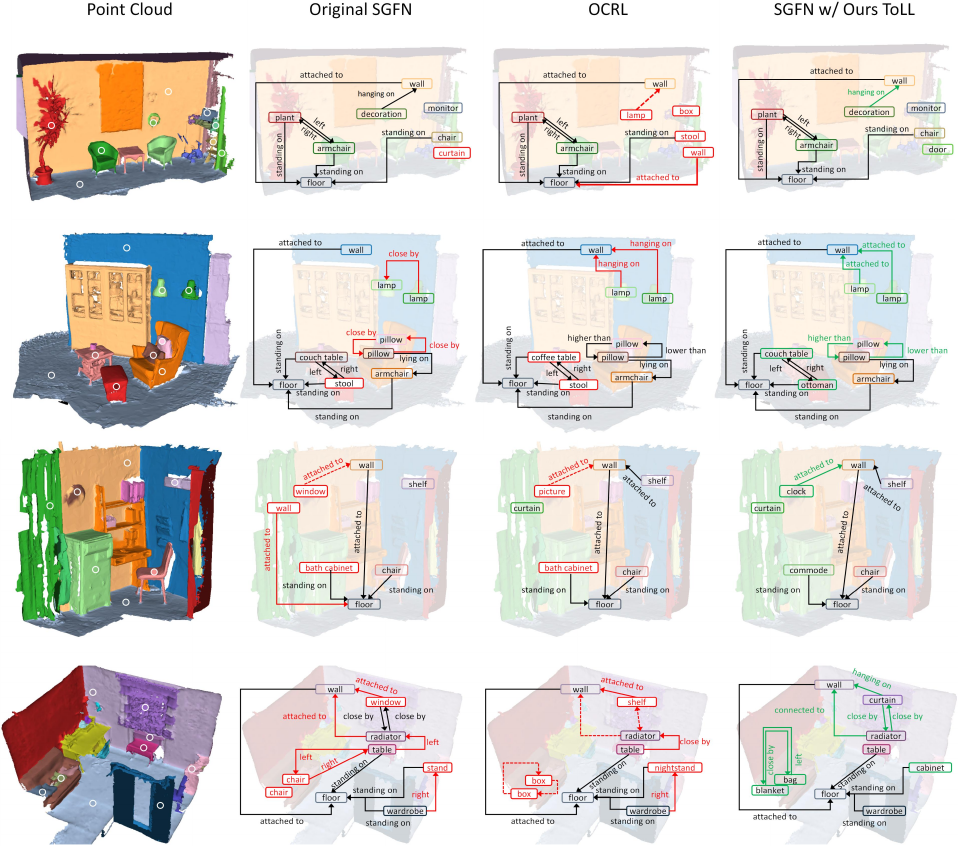}}
	\caption{Predicate A@1 for all predicate categories.}
	\label{fig5} 
\end{figure*}

\textit{Zero-Shot Generalization.} We further evaluate the capacity for zero-shot reasoning by examining performance on ``Unseen" triplets—defined as triplet configurations in the 3DSSG test set that do not appear in the training set. Our approach ``VL-SAT\textsubscript{\textit{pt}} w/ \textit{ToLL++}" demonstrates superior generalization, achieving the ``A@50" of 40.28 on unseen triplets. This substantially outperforms the from-scratch baseline by 7.41 and successfully surpasses other strong pre-training methods like MvIL (39.75), validating the powerful transferability of our learned representations.

\textbf{Qualitative Analysis.}
Figure~\ref{fig5} visualizes the generated scene graphs across four distinct indoor scenes (living rooms, bathroom, and bedroom). Our method demonstrates superior performance in two key aspects: fine-grained semantic classification and complex spatial relationship reasoning.
\subsection{Ablation Studies and Representation Analysis}
\label{sec:ablation}

We conducted extensive ablation studies on the 3DSSG. We focus on the metrics: Object ``{A@1}", Predicate ``{mA@1}" and ``{mA@3}", and Triplet ``{mA@50}" and ``{mA@100}". SGFN\textsubscript{\textit{pt}} is selected as the experimental object in this section.

\begin{table}[h]
\centering
\caption{\textbf{Ablation Studies on 3DSSG.} We analyze the impact of Anchor-Conditioned Topological Geometric Reasoning (ACTGR), Structural Multi-view Augmentation (SMA), initialization strategies, and Decoupled Layout Recovery.}
\label{tab4}
\resizebox{\linewidth}{!}{%
\begin{tabular}{l|c|cc|cc}
\toprule
\multirow{2}{*}{\textbf{Method (SGFN)/ Settings}} & \textbf{Obj} & \multicolumn{2}{c|}{\textbf{Pred}} & \multicolumn{2}{c}{\textbf{Triplet}} \\
 & A@1 & mA@1 & mA@3 & mA@50 & mA@100 \\
\midrule
\multicolumn{6}{l}{\textit{\textbf{(1) Anchor-Conditioned Topological Geometric Reasoning, ACTGR}}} \\
Baseline\textsubscript{Global Layout} & 57.62 & 47.28 & 70.20 & 60.39 & 70.85 \\
ACTGR & 58.46 & 53.87 & 75.69 & 64.92 & 73.66 \\
\midrule
\multicolumn{6}{l}{\textit{\textbf{(2) Structural Multi-view Augmentation, SMA}}} \\
SMA w/ ACTGR  & 58.68 & 54.59 & 81.36 & 66.58 & 74.32 \\
SMA w/o ACTGR & 55.36 & 50.07 & 72.64 & 61.94 & 71.63 \\
\midrule
\multicolumn{6}{l}{\textit{\textbf{(3) Object Encoder Init (with full ToLL)}}} \\
Random Initialization & 57.94 & 52.37 & 76.62 & 63.69 & 73.87 \\
PointDif Initialization & 58.68 & 54.59 & 81.36 & 66.58 & 74.32 \\
\midrule
\multicolumn{6}{l}{\textit{\textbf{(4) Decoupled Geometric Layout Recovery (with full ToLL)}}} \\
w/ Decoupled (ToLL++) & \textbf{60.64} & \textbf{56.19} & {80.79} & \textbf{67.85} & \textbf{76.25} \\
w/o Decoupled & 58.68 & 54.59 & \textbf{81.36} & 66.58 & 74.32 \\
\midrule
\multicolumn{6}{l}{\textit{\textbf{(5) Pre-training Data Scale (with full ToLL)}}} \\
Reduced (50\% ScanNet) & {56.84} & {50.85} & {76.10} & {61.25} & {72.46} \\
Default (100\% ScanNet) & 58.68 & 54.59 & {81.36} & 66.58 & 74.32\\
ScanNet \& ScanNet++ & 59.94 & 55.87 & 80.28 & 67.31 & 76.14 \\
\midrule
\multicolumn{6}{l}{\textit{\textbf{(4) Different Generative Methods (with full ToLL)}}} \\
ToLL w/ diffusion & {58.68} & {54.59} & \textbf{81.36} & {66.58} & {74.32} \\
ToLL w/ AE & 57.19 & 55.17 & 78.62 & 65.93 & 74.26 \\
ToLL++ w/ diffusion & \textbf{60.64} & \textbf{56.19} & {80.79} & \textbf{67.85} & \textbf{76.25} \\
ToLL++ w/ AE & 59.47 & 55.85 & 78.34 & 66.19 & 75.63 \\
\bottomrule
\end{tabular}%
}
\end{table}

\textbf{Effectiveness of Anchor-Conditioned Topological Reasoning (ACTGR).} We formulated a \textit{Baseline (Global Absolute Layout)} where the model is provided with the absolute spatial positions and bounding box sizes for \textbf{all} objects with 0.2 edge masking ratio. Our ACTGR retain only one random anchor's absolute attributes and mask the rest. 
As shown in Table~\ref{tab4}, the ACTGR outperforms the Global-layout Baseline, particularly in Triplet ``{mA@50}" (improvement of 4.53). Because the baseline suffers from \textbf{``shortcut learning''} where the network memorizes absolute coordinates from objects rather than learning the topological layout encoded in the edges. 

We further conduct ablation studies on different anchor ratios using both a baseline equipped solely with ACTGR and the complete ToLL. As shown in Table~\ref{actgr}, the pre-trained model achieves optimal fine-tuning performance on downstream tasks when the anchor count is exactly 1.

\begin{table}[htbp]
\caption{Ablation study of different anchor ratios.}
\label{actgr}
\centering
\large
\resizebox{\linewidth}{!}{%
\begin{tabular}{llccc} 
\toprule
\textbf{Method} & \textbf{Visible Anchors} & \textbf{Object A@1} & \textbf{Predicate mA@1} & \textbf{Triplet mA@50} \\
\midrule
ACTGR Only & Single Anchor & \textbf{58.46} & \textbf{53.87} & \textbf{64.92} \\
           & 50\% Visible Anchors & 57.83 & 46.80 & 61.30 \\
           & 100\% Visible Anchors & 57.62 & 47.28 & 60.39 \\
\midrule
ToLL (Ours) & Single Anchor & \textbf{58.68} & \textbf{54.59} & \textbf{66.58} \\
           & 50\% Visible Anchors & 58.96 & 49.85 & 62.20 \\
           & 100\% Visible Anchors & 58.49 & 47.76 & 60.65 \\
\bottomrule
\end{tabular}%
}
\end{table}

\textbf{Impact of Structural Multi-view Augmentation (SMA).} 
As shown in Table~\ref{tab4}, we analyzed the necessity of the SMA with cross-view distillation. Building upon ACTGR, by leveraging SMA-based self-distillation, we achieve effective semantic alignment while preventing feature collapse. It can be observed that ``SMA w/ ACTGR"  further improves Predicate ``mA@3" by 5.67 and Triplet ``mA@50" by 1.66 compared to the only ``ACTGR". And relying solely on SMA, ``SMA w/o ACTGR" fails to achieve superior representation learning.

\textbf{Object Encoder Initialization Strategies.} 
We evaluated two initialization protocols for the object encoder $\phi_{obj}$ when pretraining: (1) Random Initialization; (2) PointDif Initialization. 
The PointDif initialization improves Object ``A@1" by 0.74 and Predicate ``mA@1" by 2.22, compared to the random one. This indicates that leveraging the pretrained object weights could enhance the 3DSG pretraining.

\textbf{Decoupled Geometric Layout Recovery.} As shown in Table~\ref{tab4}, ToLL++ improves upon ToLL across all metrics, with a 1.96 increase in Object A@1. Decoupling shape, scale, and position enables object generation in a canonical space, preventing large objects from submerging the gradients of smaller ones and boosting geometric and semantic learning.

\textbf{Pre-training Data Scale.} As shown in Table~\ref{tab4}. We observe a positive correlation between data scale and metric outcomes: reducing the data to 50\% of ScanNet severely harms performance, while augmenting the default 100\% ScanNet dataset with 8,689 subgraph samples drawn from ScanNet++ still provides a performance boost. These results highlight the importance of large-scale pre-training for our architecture.

\textbf{Comparison of Generative Methods.} Table~\ref{tab4} compares different generative schemes. ToLL w/ diffusion consistently outperforms the AE baseline, particularly in Predicate mA@3 (81.36 vs. 78.62), demonstrating that diffusion-based recovery better captures complex spatial layouts. Notably, ToLL++ w/ diffusion achieves state-of-the-art results, reaching 60.64 in Object A@1 and 56.19 in Predicate mA@1. 

\begin{figure}[h]
	\centering
	\subfloat{
		\includegraphics[width=0.98\linewidth]{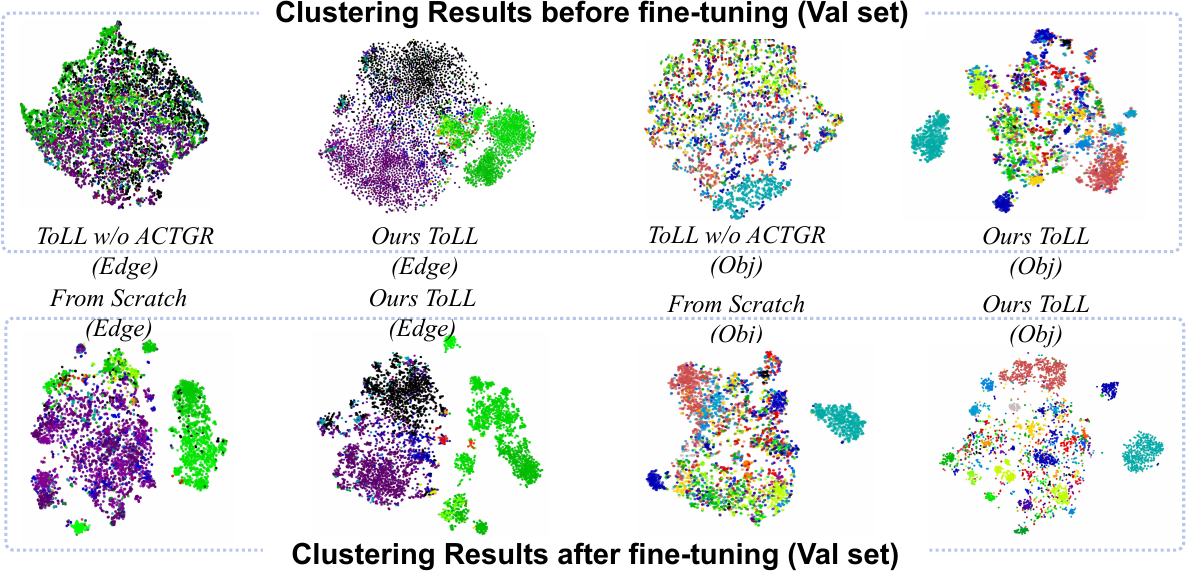}}
	\caption{Visualization of latent features clustering.}
	\label{fig6} 
\end{figure}

\textbf{Cluster Quality Evaluation.}
As shown in Figure~\ref{fig6}, compared to the weight from ``\textit{ToLL w/o ACTGR}", the full ToLL demonstrates that the information bottleneck constructed by ACTGR compels the model to learn topological edge structures, focusing more on the semantic and spatial information of the entire graph. Moreover, the fine-tuning analysis on val set confirms that our method generates more distinct clusters than training from scratch.

\begin{figure}[h]
	\centering
	\subfloat{
		\includegraphics[width=0.98\linewidth]{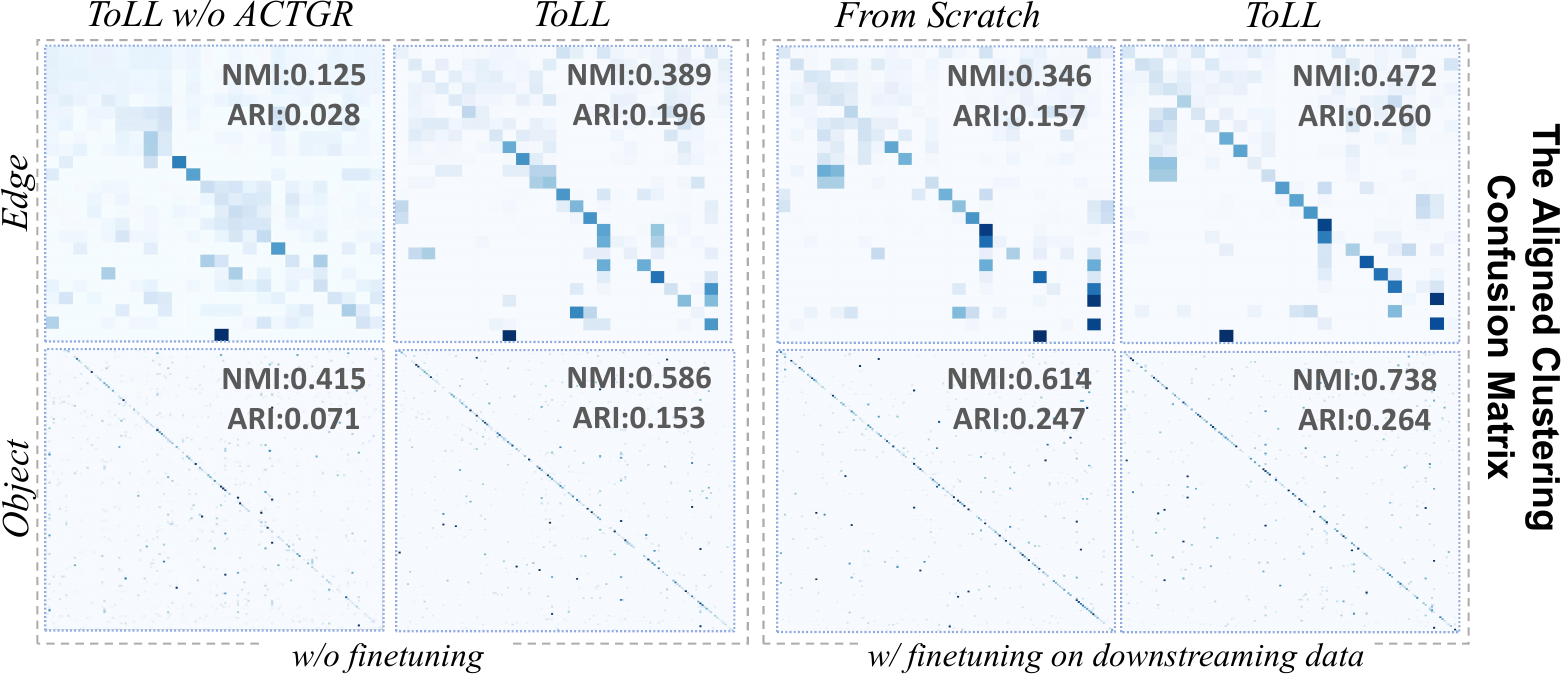}}
	\caption{Visualization of clustering confusion matrix.}
	\label{fig7} 
\end{figure}

Figure~\ref{fig7} is the clustering confusion matrix on the 3DSSG Val. Set. The diagonal elements indicate the clustering consistency with GT labels after optimal matching. We report the Normalized Mutual Information (NMI) and Adjusted Rand Index (ARI) of the node and edge features among weights with ``\textit{ToLL w/o ACTGR}", ``\textit{ToLL}" and ``\textit{From Scratch}". The first two methods involve no fine-tuning, whereas ``\textit{From Scratch}" represents supervised learning on 3DSSG with random initialization. Notably, ``\textit{ToLL}" outperforms the supervised one in predicate-level NMI and ARI metrics. This demonstrates that our pre-training scheme achieves superior semantic clustering.

\begin{figure}[t!]
	\centering
	\subfloat{
		\includegraphics[width=0.98\linewidth]{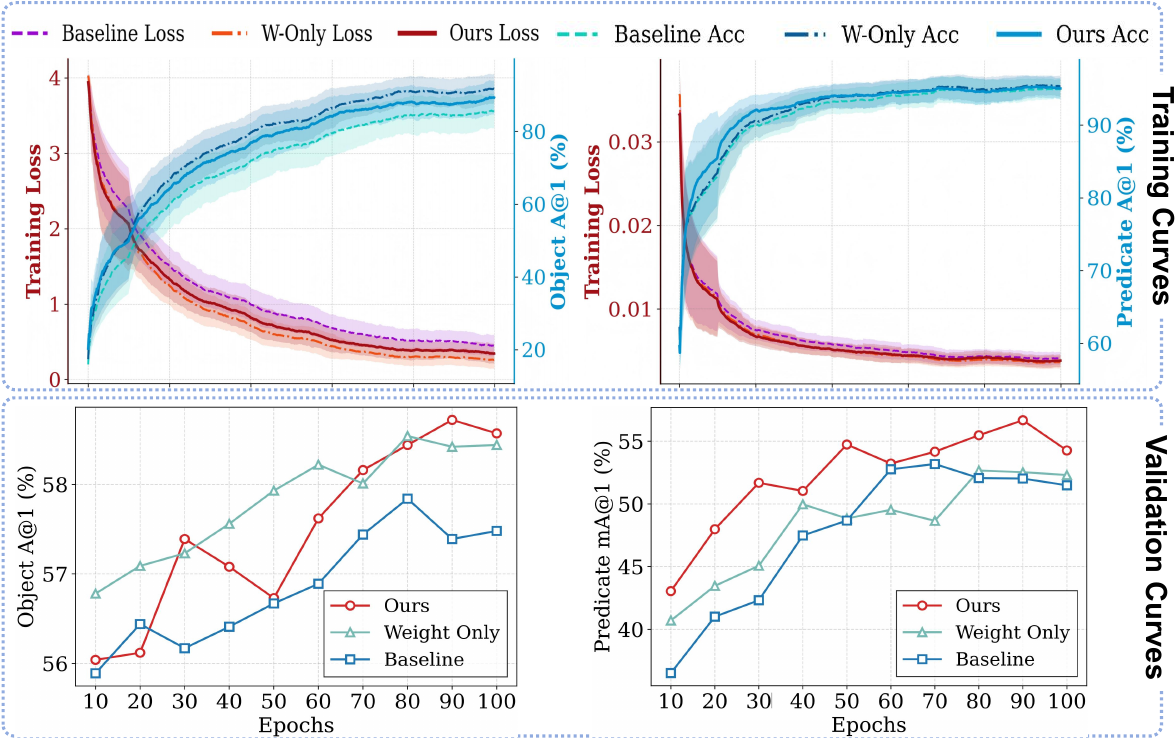}}
	\caption{Analysis of accuracy curves (Train and Validation).}
	\label{fig8} 
\end{figure}

\textbf{Analysis of Accuracy Curves.} As depicted in Figure~\ref{fig8}, Our ToLL results in faster convergence compared to the baseline. The predicate training accuracy exhibits a steeper ascent than the baseline and the object weights initialization with \textit{PointDif} 'W-only Acc'. During evaluation, our scheme outperforms all comparisons, securing the highest scores on both Object ``A@1" and Predicate ``mA@1".

\begin{figure}[h]
	\centering
	\subfloat{
		\includegraphics[width=0.98\linewidth]{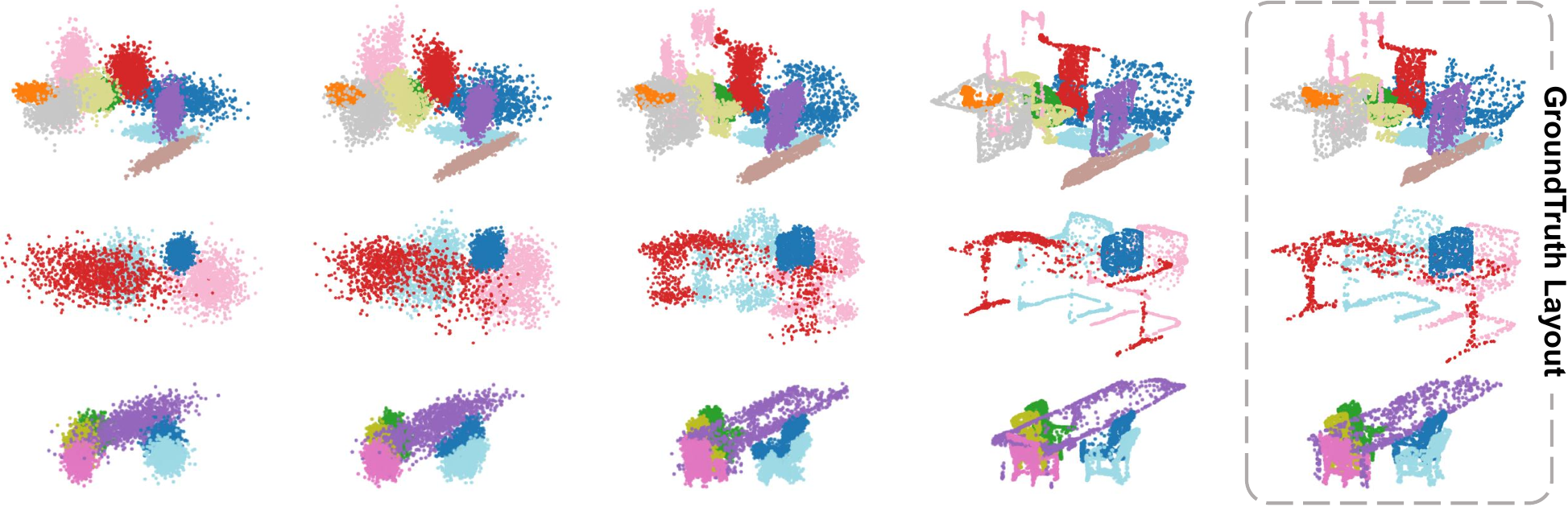}}
\caption{Visualization of diffusion layout recovery.}
	\label{fig9} 
\end{figure}

\textbf{Layout Recovery Results.} Figure~\ref{fig9} illustrates the results in  our topological layout learning task. With the progression of diffusion iterations, our scheme leverages semantic conditional features to reconstruct scene point clouds characterized by fine geometric details and accurate spatial layouts.

\section{Conclusion and Future Work}

This paper presents a new Topological Layout Learning with asymmetric cross-view distillation framework for the pre-training of 3DSG generation. By integrating Anchor-Conditioned Topological Geometric Reasoning with the self-distillation on Structural Multi-view Augmentation, it turns a layout recovery task into a robust proxy for learning topological and semantic dependencies. Extensive experiments show that this method could achieve state-of-the-art performance on 3DSSG benchmark. Nevertheless, some challenges still exist, due to the inherent sparsity of data and the high noise levels in relational triplet data. To address these bottlenecks, future work will focus on the construction of better large-scale high-quality 3DSG datasets, and then solve the ambiguous definitions of predicates.

\vfill

\bibliographystyle{IEEEtran}
\bibliography{IEEEabrv,IEEEexample}

@article{you2020graph,
  title={Graph contrastive learning with augmentations},
  author={You, Yuning and Chen, Tianlong and Sui, Yongduo and Chen, Ting and Wang, Zhangyang and Shen, Yang},
  journal={Advances in neural information processing systems},
  volume={33},
  pages={5812--5823},
  year={2020}
}

@article{wei20233d,
  title={3d scene graph generation from point clouds},
  author={Wei, Wenwen and Wei, Ping and Qin, Jialu and Liao, Zhimin and Wang, Shuaijie and Cheng, Xiang and Liu, Meiqin and Zheng, Nanning},
  journal={IEEE Transactions on Multimedia},
  volume={26},
  pages={5358--5368},
  year={2023},
  publisher={IEEE}
}

@article{wang2024weakly,
  title={Weakly-supervised 3d scene graph generation via visual-linguistic assisted pseudo-labeling},
  author={Wang, Xu and Li, Yifan and Zhang, Qiudan and Wu, Wenhui and Li, Mark Junjie and Ma, Lin and Jiang, Jianmin},
  journal={IEEE Transactions on Multimedia},
  volume={26},
  pages={11164--11175},
  year={2024},
  publisher={IEEE}
}

@inproceedings{zhu2021graph,
  title={Graph contrastive learning with adaptive augmentation},
  author={Zhu, Yanqiao and Xu, Yichen and Yu, Feng and Liu, Qiang and Wu, Shu and Wang, Liang},
  booktitle={Proceedings of the web conference 2021},
  pages={2069--2080},
  year={2021}
}

@inproceedings{hou2022graphmae,
  title={Graphmae: Self-supervised masked graph autoencoders},
  author={Hou, Zhenyu and Liu, Xiao and Cen, Yukuo and Dong, Yuxiao and Yang, Hongxia and Wang, Chunjie and Tang, Jie},
  booktitle={Proceedings of the 28th ACM SIGKDD conference on knowledge discovery and data mining},
  pages={594--604},
  year={2022}
}

@inproceedings{kingmaauto,
  title={Auto-encoding variational Bayes},
  author={Kingma, Diederik P and Welling, Max},
  booktitle={Int. Conf. on Learning Representations},
  year={2014}
}

@article{van2017neural,
  title={Neural discrete representation learning},
  author={Van Den Oord, Aaron and Vinyals, Oriol and others},
  journal={Advances in neural information processing systems},
  volume={30},
  year={2017}
}

@article{van2025joint,
  title={Joint Embedding vs Reconstruction: Provable Benefits of Latent Space Prediction for Self Supervised Learning},
  author={Van Assel, Hugues and Ibrahim, Mark and Biancalani, Tommaso and Regev, Aviv and Balestriero, Randall},
  journal={arXiv preprint arXiv:2505.12477},
  year={2025}
}

@inproceedings{wusimplifying,
  title={Simplifying DINO via Coding Rate Regularization},
  author={Wu, Ziyang and Zhang, Jingyuan and Pai, Druv and Wang, XuDong and Singh, Chandan and Yang, Jianwei and Gao, Jianfeng and Ma, Yi},
  booktitle={Forty-second International Conference on Machine Learning},
  year={2025}
}

@article{caron2020unsupervised,
  title={Unsupervised learning of visual features by contrasting cluster assignments},
  author={Caron, Mathilde and Misra, Ishan and Mairal, Julien and Goyal, Priya and Bojanowski, Piotr and Joulin, Armand},
  journal={Advances in neural information processing systems},
  volume={33},
  pages={9912--9924},
  year={2020}
}

@inproceedings{caron2021emerging,
  title={Emerging properties in self-supervised vision transformers},
  author={Caron, Mathilde and Touvron, Hugo and Misra, Ishan and J{\'e}gou, Herv{\'e} and Mairal, Julien and Bojanowski, Piotr and Joulin, Armand},
  booktitle={Proceedings of the IEEE/CVF international conference on computer vision},
  pages={9650--9660},
  year={2021}
}

@inproceedings{anderson2018vision,
  title={Vision-and-language navigation: Interpreting visually-grounded navigation instructions in real environments},
  author={Anderson, Peter and Wu, Qi and Teney, Damien and Bruce, Jake and Johnson, Mark and S{\"u}nderhauf, Niko and Reid, Ian and Gould, Stephen and Van Den Hengel, Anton},
  booktitle={Proceedings of the IEEE conference on computer vision and pattern recognition},
  pages={3674--3683},
  year={2018}
}

@inproceedings{zitkovich2023rt,
  title={Rt-2: Vision-language-action models transfer web knowledge to robotic control},
  author={Zitkovich, Brianna and Yu, Tianhe and Xu, Sichun and Xu, Peng and Xiao, Ted and Xia, Fei and Wu, Jialin and Wohlhart, Paul and Welker, Stefan and Wahid, Ayzaan and others},
  booktitle={Conference on Robot Learning},
  pages={2165--2183},
  year={2023},
  organization={PMLR}
}

@inproceedings{wald2020learning,
  title={Learning 3d semantic scene graphs from 3d indoor reconstructions},
  author={Wald, Johanna and Dhamo, Helisa and Navab, Nassir and Tombari, Federico},
  booktitle={Proceedings of the IEEE/CVF Conference on Computer Vision and Pattern Recognition},
  pages={3961--3970},
  year={2020}
}

@article{zhang2021knowledge,
  title={Knowledge-inspired 3d scene graph prediction in point cloud},
  author={Zhang, Shoulong and Hao, Aimin and Qin, Hong and others},
  journal={Advances in Neural Information Processing Systems},
  volume={34},
  pages={18620--18632},
  year={2021}
}

@inproceedings{wang2023vl,
  title={Vl-sat: Visual-linguistic semantics assisted training for 3d semantic scene graph prediction in point cloud},
  author={Wang, Ziqin and Cheng, Bowen and Zhao, Lichen and Xu, Dong and Tang, Yang and Sheng, Lu},
  booktitle={Proceedings of the IEEE/CVF conference on computer vision and pattern recognition},
  pages={21560--21569},
  year={2023}
}

@inproceedings{chen2024clip,
  title={Clip-driven open-vocabulary 3d scene graph generation via cross-modality contrastive learning},
  author={Chen, Lianggangxu and Wang, Xuejiao and Lu, Jiale and Lin, Shaohui and Wang, Changbo and He, Gaoqi},
  booktitle={Proceedings of the IEEE/CVF Conference on Computer Vision and Pattern Recognition},
  pages={27863--27873},
  year={2024}
}

@inproceedings{koch2024lang3dsg,
  title={Lang3dsg: Language-based contrastive pre-training for 3d scene graph prediction},
  author={Koch, Sebastian and Hermosilla, Pedro and Vaskevicius, Narunas and Colosi, Mirco and Ropinski, Timo},
  booktitle={2024 International Conference on 3D Vision (3DV)},
  pages={1037--1047},
  year={2024},
  organization={IEEE}
}

@inproceedings{zhang2021exploiting,
  title={Exploiting edge-oriented reasoning for 3d point-based scene graph analysis},
  author={Zhang, Chaoyi and Yu, Jianhui and Song, Yang and Cai, Weidong},
  booktitle={Proceedings of the IEEE/CVF conference on computer vision and pattern recognition},
  pages={9705--9715},
  year={2021}
}

@inproceedings{heoobject,
  title={Object-Centric Representation Learning for Enhanced 3D Semantic Scene Graph Prediction},
  author={Heo, KunHo and Kim, GiHyeon and Kim, SuYeon and Cho, MyeongAh},
  booktitle={The Thirty-ninth Annual Conference on Neural Information Processing Systems},
  year={2025}

}

@inproceedings{wu2023incremental,
  title={Incremental 3d semantic scene graph prediction from rgb sequences},
  author={Wu, Shun-Cheng and Tateno, Keisuke and Navab, Nassir and Tombari, Federico},
  booktitle={Proceedings of the IEEE/CVF conference on computer vision and pattern recognition},
  pages={5064--5074},
  year={2023}
}

@article{feng2025hyperrectangle,
  title={Hyperrectangle embedding for debiased 3D scene graph prediction from RGB sequences},
  author={Feng, Mingtao and Yan, Chenbo and Wu, Zijie and Dong, Weisheng and Wang, Yaonan and Mian, Ajmal},
  journal={IEEE Transactions on Pattern Analysis and Machine Intelligence},
  year={2025},
  publisher={IEEE}
}

@inproceedings{xie2020pointcontrast,
  title={Pointcontrast: Unsupervised pre-training for 3d point cloud understanding},
  author={Xie, Saining and Gu, Jiatao and Guo, Demi and Qi, Charles R and Guibas, Leonidas and Litany, Or},
  booktitle={European conference on computer vision},
  pages={574--591},
  year={2020},
  organization={Springer}
}

@inproceedings{zhang2021self,
  title={Self-supervised pretraining of 3d features on any point-cloud},
  author={Zhang, Zaiwei and Girdhar, Rohit and Joulin, Armand and Misra, Ishan},
  booktitle={Proceedings of the IEEE/CVF international conference on computer vision},
  pages={10252--10263},
  year={2021}
}

@inproceedings{wu2025sonata,
  title={Sonata: Self-supervised learning of reliable point representations},
  author={Wu, Xiaoyang and DeTone, Daniel and Frost, Duncan and Shen, Tianwei and Xie, Chris and Yang, Nan and Engel, Jakob and Newcombe, Richard and Zhao, Hengshuang and Straub, Julian},
  booktitle={Proceedings of the Computer Vision and Pattern Recognition Conference},
  pages={22193--22204},
  year={2025}
}

@article{pang2023masked,
  title={Masked autoencoders for 3d point cloud self-supervised learning},
  author={Pang, Yatian and Tay, Eng Hock Francis and Yuan, Li and Chen, Zhenghua},
  journal={World Scientific Annual Review of Artificial Intelligence},
  volume={1},
  pages={2440001},
  year={2023},
  publisher={World Scientific}
}

@inproceedings{zheng2024point,
  title={Point cloud pre-training with diffusion models},
  author={Zheng, Xiao and Huang, Xiaoshui and Mei, Guofeng and Hou, Yuenan and Lyu, Zhaoyang and Dai, Bo and Ouyang, Wanli and Gong, Yongshun},
  booktitle={Proceedings of the IEEE/CVF Conference on Computer Vision and Pattern Recognition},
  pages={22935--22945},
  year={2024}
}

@inproceedings{koch2024sgrec3d,
  title={Sgrec3d: Self-supervised 3d scene graph learning via object-level scene reconstruction},
  author={Koch, Sebastian and Hermosilla, Pedro and Vaskevicius, Narunas and Colosi, Mirco and Ropinski, Timo},
  booktitle={Proceedings of the IEEE/CVF Winter Conference on Applications of Computer Vision},
  pages={3404--3414},
  year={2024}
}

@inproceedings{ma2024heterogeneous,
  title={Heterogeneous graph learning for scene graph prediction in 3d point clouds},
  author={Ma, Yanni and Liu, Hao and Pei, Yun and Guo, Yulan},
  booktitle={European Conference on Computer Vision},
  pages={274--291},
  year={2024},
  organization={Springer}
}

@inproceedings{feng20233d,
  title={3D spatial multimodal knowledge accumulation for scene graph prediction in point cloud},
  author={Feng, Mingtao and Hou, Haoran and Zhang, Liang and Wu, Zijie and Guo, Yulan and Mian, Ajmal},
  booktitle={Proceedings of the IEEE/CVF Conference on Computer Vision and Pattern Recognition},
  pages={9182--9191},
  year={2023}
}

@inproceedings{zhangconcerto,
  title={Concerto: Joint 2D-3D Self-Supervised Learning Emerges Spatial Representations},
  author={Zhang, Yujia and Wu, Xiaoyang and Lao, Yixing and Wang, Chengyao and Tian, Zhuotao and Wang, Naiyan and Zhao, Hengshuang},
  booktitle={The Thirty-ninth Annual Conference on Neural Information Processing Systems}
}

@inproceedings{zhouimage,
  title={Image BERT Pre-training with Online Tokenizer},
  author={Zhou, Jinghao and Wei, Chen and Wang, Huiyu and Shen, Wei and Xie, Cihang and Yuille, Alan and Kong, Tao},
  booktitle={International Conference on Learning Representations}
}

@inproceedings{you2021graph,
  title={Graph contrastive learning automated},
  author={You, Yuning and Chen, Tianlong and Shen, Yang and Wang, Zhangyang},
  booktitle={International conference on machine learning},
  pages={12121--12132},
  year={2021},
  organization={PMLR}
}

@article{xu2021infogcl,
  title={Infogcl: Information-aware graph contrastive learning},
  author={Xu, Dongkuan and Cheng, Wei and Luo, Dongsheng and Chen, Haifeng and Zhang, Xiang},
  journal={Advances in Neural Information Processing Systems},
  volume={34},
  pages={30414--30425},
  year={2021}
}

@inproceedings{tian2023heterogeneous,
  title={Heterogeneous graph masked autoencoders},
  author={Tian, Yijun and Dong, Kaiwen and Zhang, Chunhui and Zhang, Chuxu and Chawla, Nitesh V},
  booktitle={Proceedings of the AAAI conference on artificial intelligence},
  volume={37},
  number={8},
  pages={9997--10005},
  year={2023}
}

@book{hamilton2020graph,
  title={Graph representation learning},
  author={Hamilton, William L},
  year={2020},
  publisher={Morgan \& Claypool Publishers}
}

@article{hu2024survey,
  title={A survey on information bottleneck},
  author={Hu, Shizhe and Lou, Zhengzheng and Yan, Xiaoqiang and Ye, Yangdong},
  journal={IEEE Transactions on Pattern Analysis and Machine Intelligence},
  volume={46},
  number={8},
  pages={5325--5344},
  year={2024},
  publisher={IEEE}
}

@inproceedings{armeni20193d,
  title={3d scene graph: A structure for unified semantics, 3d space, and camera},
  author={Armeni, Iro and He, Zhi-Yang and Gwak, JunYoung and Zamir, Amir R and Fischer, Martin and Malik, Jitendra and Savarese, Silvio},
  booktitle={Proceedings of the IEEE/CVF international conference on computer vision},
  pages={5664--5673},
  year={2019}
}

@article{liu2022graph,
  title={Graph self-supervised learning: A survey},
  author={Liu, Yixin and Jin, Ming and Pan, Shirui and Zhou, Chuan and Zheng, Yu and Xia, Feng and Yu, Philip S},
  journal={IEEE transactions on knowledge and data engineering},
  volume={35},
  number={6},
  pages={5879--5900},
  year={2022},
  publisher={IEEE}
}

@article{wu2021self,
  title={Self-supervised learning on graphs: Contrastive, generative, or predictive},
  author={Wu, Lirong and Lin, Haitao and Tan, Cheng and Gao, Zhangyang and Li, Stan Z},
  journal={IEEE Transactions on Knowledge and Data Engineering},
  volume={35},
  number={4},
  pages={4216--4235},
  year={2021},
  publisher={IEEE}
}

@inproceedings{huang2026multi,
  title={Multi-view Invariance Learning for 3D Scene Graph Pre-training via Collaborative Cross-Modal Regularization},
  author={Huang, Yucheng and Ji, Luping and Xiao, Ruijie and Sun, Jiayuan},
  booktitle={Proceedings of the AAAI Conference on Artificial Intelligence},
  volume={40},
  number={7},
  pages={5203--5211},
  year={2026}
}

\end{document}